\documentclass[11pt]{article}

\usepackage[preprint]{acl}

\usepackage{microtype}

\usepackage{times}
\usepackage{latexsym}
\usepackage[T1]{fontenc}
\usepackage[utf8]{inputenc}

\usepackage{graphicx}
\usepackage{booktabs}

\title{Yo-ByT5: Efficient and High-Fidelity Diacritic Restoration for Yor\`ub\'a}

\author{
  \textbf{Ahmad Samuel Gali}$^{1}$ \quad
  \textbf{Shamsuddeen Hassan Muhammad}$^{2,3}$ \\[1.5ex]
  $^{1}$University of Lagos \\
  $^{2}$Bayero University Kano \quad
  $^{3}$Imperial College London 
}

\begin{document}
\maketitle

\begin{abstract}
Yor\`ub\'a is a widely spoken tonal language that depends on diacritics to avoid lexical ambiguity. However, it is often written without these diacritics, thereby hindering downstream Natural Language Processing (NLP) tasks. In this paper, we introduce Yo-ByT5, a byte-level Automatic Diacritic Restoration (ADR) model fine-tuned from ByT5-small. We evaluate Yo-ByT5 alongside five publicly released Yor\`ub\'a ADR models and one open-weight large language model (LLM) on the YAD benchmark under a consistent protocol. Our results demonstrate that Yo-ByT5 matches the performance of the strongest existing model, mT5-base, with a DER of 10.14\% and a CER of 3.48\%. Furthermore, it exhibits superior text fidelity despite using approximately half the parameter count of mT5-base. We also release our training code and model outputs, as well as call for the development of a larger, purpose-built benchmark for Yor\`ub\'a diacritic restoration.

\end{abstract}

\section{Introduction}

\subsection{The problem}

The Yor\`ub\'a language is a Niger-Congo language and one of the three most spoken languages in Nigeria, with more than 50 million speakers across the country and other parts of Africa \citep{inuwa-dutse-2025}. However, despite its mass adoption globally, Yor\`ub\'a is classified as a low-resource language. The low availability of computational linguistic resources necessitates this description, despite the rich cultural heritage of the language. The field of ADR makes this resource scarcity particularly evident, as diacritics affect almost every downstream task, from text-to-speech to machine translation and corpus annotation. A systematic review of 105 studies across ten years highlights diacritic dependency as one of the central unsolved challenges for Yor\`ub\'a NLP \citep{jimoh-etal-2025}.

This challenge stems from ambiguity in the Yor\`ub\'a writing system. Yor\`ub\'a employs diacritics---tone marks and underdots---to disambiguate text, reducing the need for context and auxiliary information to decipher meaning. For instance, in a Yor\`ub\'a corpus of about one million words, 85\% of words carry diacritics, and 32\% of unmarked word types have two or more valid diacritised counterparts \citep{orife-2018}. Table~\ref{tab:ambiguity} shows an example: the word \emph{igba} has four valid diacritised readings. However, in casual settings such as social media posts and comments, speakers often write in the plain Latin alphabet and rely on context for disambiguation. For downstream NLP tasks, this plain text representation severely degrades performance, necessitating automatic diacritic restoration models.

\begin{table}[t]
  \centering
  \small
  \caption{One undiacritised Yor\`ub\'a form, \emph{igba}, and four valid readings.}
  \label{tab:ambiguity}
  \begin{tabular}{@{}llll@{}}
    \toprule
    \textbf{Undiacritised} & \textbf{Reading} & \textbf{Tone pattern} & \textbf{Meaning} \\
    \midrule
    igba & igb\'a  & mid--high & calabash \\
         & \`igb\`a & low--low  & time, period \\
         & \`igb\'a & low--high & garden egg \\
         & igba    & mid--mid  & two hundred \\
    \bottomrule
  \end{tabular}
\end{table}

\subsection{Modelling choices for the task}

ADR is a specialised task with various methodological approaches, ranging from rule- and syllable-based techniques \citep{adegbola-odilinye-2012, asahiah-etal-2017} to modern neural network architectures \citep{naplava-etal-2018, olawole-etal-2024, orife-etal-2020}. One standard approach frames ADR as a machine translation task using sequence-to-sequence neural models \citep{orife-2018}. This framing is intuitive, as diacritics can be treated as properties restored to base characters. However, byte-level modelling advances this concept further.

A byte-level model processes raw text directly without learning a fixed subword vocabulary. This property makes byte-level models robust to spelling variations and eliminates out-of-vocabulary issues \citep{xue-etal-2022}. Furthermore, they can process text in any language inherently and simplify technical pipelines by removing complex preprocessing steps that could introduce errors. Additionally, subword-based models like mT5 allocate a large share of their parameters (approximately 66\% in mT5-base) to vocabulary embeddings and output matrices. In a byte-level model, these parameters can be reallocated to increase network depth or width \citep{xue-etal-2021, xue-etal-2022}.

\subsection{This paper}

In this paper, we introduce Yo-ByT5, a byte-level model based on ByT5-small, fine-tuned for ADR in Yor\`ub\'a text. We evaluate Yo-ByT5 alongside five other publicly released Yor\`ub\'a diacritisers and one open-weight language model across four core metrics: Character Error Rate (CER), Word Error Rate (WER), Diacritic Error Rate (DER), and Word Diacritic Error Rate (WDER). In addition, we analyse decoding strategies in the inference pipeline and assess text fidelity by isolating diacritic errors from structural text modifications. Finally, we release our code and per-sentence outputs at \url{https://github.com/lazy-monster/yo-byt5} to enable reproducible comparisons and extensions by the research community.

\subsection{Contributions}

In this paper, we make the following contributions:

\begin{enumerate}
  \item Yo-ByT5, an open-weight byte-level ADR model for Yor\`ub\'a, alongside its training procedure.
  \item A comparison of seven models for the diacritic restoration task including Yo-ByT5, one open-weight language model, and five other publicly available diacritisers on the YAD benchmark under the same scoring protocol.
  \item An error typology analysis that demonstrates that text fidelity, rather than error rate, most sharply separates these models from one another, and that Yo-ByT5 preserves the input text far more reliably than any comparably accurate model.
  \item Released code, model outputs, and a small patch for the YAD benchmark discovered during evaluation to enable reproducible benchmark results.
  \item A call for the development of a larger, purpose-built benchmark dataset for Yor\`ub\'a diacritic restoration.
\end{enumerate}

\section{Related Work}

\subsection{Yor\`ub\'a diacritic restoration}

Early neural work in diacritic restoration by \citet{orife-2018} framed the task as sequence-to-sequence machine translation using attentive recurrent networks, releasing models and data to the public. While their training relied primarily on biblical text, subsequent work identified generalisation limitations on colloquial text, contractions, and loan words \citep{orife-etal-2020}. Soft-attention sequence-to-sequence models exhibited reduced performance on informal linguistic variants. To address this, \citet{orife-etal-2020} constructed a larger multi-domain corpus, incorporating the JW300 Yor\`ub\'a dataset (11.5M words out of 13.9M total), and introduced a news test set for evaluation on modern, non-literary text.

Another key contribution is the YAD benchmark, used for evaluation in this paper. Several multilingual T5 models and a Yor\`ub\'a-specific T5 model were fine-tuned on YAD's training split and evaluated using BLEU and ChrF metrics \citep{olawole-etal-2024}. More recently, \citet{ayoola-olukanmi-2026} benchmarked four diacritisation models, including general LLMs and fine-tuned models such as mT5-base, reporting CER and DER metrics under their protocol.

\subsection{Corpora and benchmark data}

MENYO-20k is a multi-domain parallel corpus for Yor\`ub\'a and English, constructed specifically with complete and accurate diacritisation, making it highly suitable for ADR tasks \citep{adelani-etal-2021}. It serves as a foundational source for the YAD benchmark alongside JW300 and the Yor\`ub\'a Bible, as well as the training data for Yo-ByT5. JW300 \citep{agic-vulic-2019} remains a prominent resource, though NLP research for Nigerian languages often relies on repurposing existing resources rather than constructing new, high-quality annotated datasets \citep{inuwa-dutse-2025}.

\subsection{Model architectures}

mT5 is a multilingual variant of T5 pre-trained on a Common Crawl dataset spanning 101 languages. The base variant (582.4M parameters) serves as one of the baseline architectures evaluated in this study \citep{xue-etal-2021}. ByT5 replaces subword vocabularies with raw UTF-8 bytes, offering increased robustness to noise and character variations \citep{xue-etal-2022}. Our proposed model, Yo-ByT5, fine-tunes ByT5-small (approximately 300M parameters) for Yor\`ub\'a ADR.

\subsection{Large language models for diacritisation}

Large language models (LLMs) have demonstrated broad capability across NLP tasks, raising questions regarding their efficacy in ADR. \citet{nadas-diosan-2025} evaluated LLMs on Romanian diacritic restoration, finding that performance varied significantly across model families, with several performing below an echo baseline and highlighting the continued necessity of specialised fine-tuning. \citet{toyin-etal-2025} evaluated 12 LLMs against specialised diacritisers on Arabic and Yor\`ub\'a using the MultiDiac test set. While several LLMs outperformed specialised baselines, smaller models frequently exhibited hallucination and text degradation. In this study, we benchmark specialised Yor\`ub\'a diacritisers, including Yo-ByT5, against a 120B parameter open-weight LLM on the YAD benchmark.

\section{Yo-ByT5}

\subsection{Architecture and rationale}

The underlying architecture for Yo-ByT5 is \texttt{google/byt5-small}, a pre-trained model with 299,637,760 parameters. A byte-level model was selected because diacritisation is fundamentally a character- and byte-level operation involving combining marks attached to base characters. Yo-ByT5 restores both tone marks and underdots to unmarked Yor\`ub\'a text.

\subsection{Training data}

\begin{figure*}[t]
  \centering
  \includegraphics[width=\textwidth]{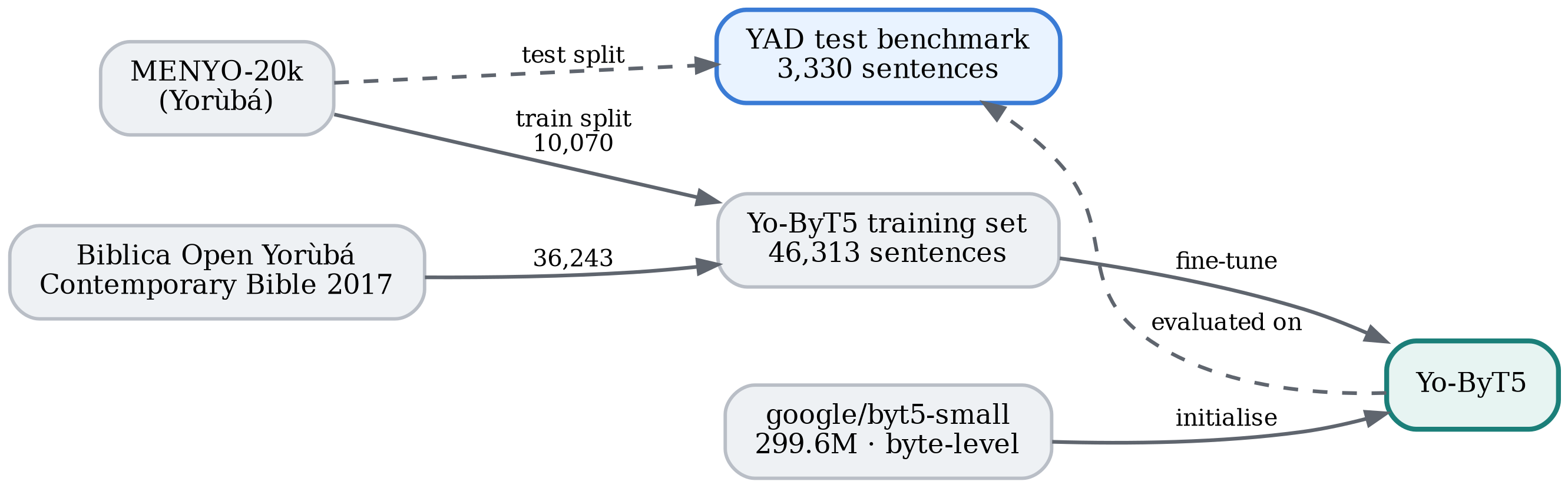}
  \caption{Yo-ByT5 corpus and fine-tuning lineage. The MENYO-20k train split and the Biblica Open Yor\`ub\'a Contemporary Bible 2017 form the training set; the model is initialised from \texttt{google/byt5-small} and fine-tuned on it. The YAD test benchmark, on which Yo-ByT5 is evaluated, is drawn from the MENYO-20k test split.}
  \label{fig:lineage}
\end{figure*}

The Yo-ByT5 training set combines the Yor\`ub\'a training split of MENYO-20k (10,070 sentences) and the Biblica Open Yor\`ub\'a Contemporary Bible 2017 (36,243 segments), for a total of 46,313 training sentences. We hold out a further 5,305 sentences for validation and 6,633 for internal testing (Figure~\ref{fig:lineage}).

\subsection{Training procedure}

We fine-tuned \texttt{google/byt5-small} on a Google Cloud TPU v6e-8 in two phases. Both phases used AdamW with linear decay, weight decay 0.01, and gradient clipping at a norm of 0.5. The per-device batch size was 4 with gradient accumulation 2 across 8 TPU cores, for an effective global batch size of 64. Both phases tokenised source and target text to at most 1024 bytes and ran for 4 epochs on the training set. The first phase used a learning rate of 2e-4 with 300 warmup steps. The second phase started from the first-phase checkpoint, used a learning rate of 1e-4 with no warmup, and produced the released Yo-ByT5 checkpoint.

\subsection{Contamination check}

We audit the internal training data against the dev and test splits of the YAD benchmark. The only overlaps are two short strings that also occur in YAD's official train split: the greeting \emph{\d{E} \d{s}eun} (``thank you''), in 4 of the 3,330 test sentences and 11 dev sentences, and a boilerplate header line, in 8 dev sentences. Therefore, we can evaluate Yo-ByT5 on the YAD benchmark without meaningful leakage.

\section{Experimental Setup}

\subsection{Benchmark}

The benchmark we evaluate the models on is the test split of the YAD dataset, which contains 3,330 sentences drawn from MENYO-20k \citep{olawole-etal-2024}.

\subsection{Models compared}

We compare seven models in total: Yo-ByT5; five publicly released Yor\`ub\'a ADR models (mT5-base, two omowe-T5 variants, ByT5-small menyo, and mT5-small); and one open-weight LLM baseline (gpt-oss:120b evaluated via Ollama Cloud with one-shot prompting at temperature 0). Table~\ref{tab:models} lists each model with its public repository, architecture, and parameter count.

\begin{table*}[t]
  \centering
  \small
  \caption{Evaluated models: Yo-ByT5, five publicly released Yor\`ub\'a diacritisers, and one open-weight LLM baseline.}
  \label{tab:models}
  \begin{tabular}{@{}lllr@{}}
    \toprule
    \textbf{Model} & \textbf{Repository / provider} & \textbf{Architecture} & \textbf{Params} \\
    \midrule
    Yo-ByT5          & \texttt{lazymonster/yobyt5-restoration}                 & ByT5, byte-level  & 299.6M \\
    mT5-base         & \texttt{Davlan/mT5\_base\_yoruba\_adr}                  & mT5, subword      & 582.4M \\
    omowe-T5 all-und & \texttt{Davlan/omowe-t5-small-diacritizer-all-und-full} & T5-small, subword & 76.9M \\
    omowe-T5 menyo   & \texttt{Davlan/omowe-t5-small-diacritizer-menyo}        & T5-small, subword & 76.9M \\
    ByT5-small menyo & \texttt{Davlan/byt5-small-diacritizer-menyo}            & ByT5, byte-level  & 299.6M \\
    mT5-small menyo  & \texttt{Davlan/mt5-small-diacritizer-menyo}             & mT5, subword      & 300.2M \\
    gpt-oss:120b     & Ollama Cloud (one-shot)                                & decoder-only LLM  & $\approx$120B \\
    \bottomrule
  \end{tabular}
\end{table*}

\subsection{Decoding}

We use deterministic decoding throughout the evaluation pipeline for reproducibility. Furthermore, since the observed performance of weaker models changes by up to 10.5 CER points when decoding strategy is varied, we use both greedy and beam decoding for comprehensiveness. However, we find that the two strongest models change by at most 0.12 points regardless of the decoding strategy.

\subsection{Metrics}

We report eight evaluation metrics: Word Error Rate (WER), Character Error Rate (CER), Diacritic Error Rate (DER), Word Diacritic Error Rate (WDER), Underdot DER (DER-underdot), Tone DER (DER-tone), BLEU, and ChrF.

WER and CER measure edit distance (substitutions, deletions, insertions) between model outputs and reference text at the word and character levels using Levenshtein distance \citep{levenshtein-1966, ayoola-olukanmi-2026}. DER measures the percentage of diacritised reference letters assigned incorrect diacritics \citep{zitouni-etal-2006, fadel-etal-2019, ayoola-olukanmi-2026}. We also break down DER into DER-underdot and DER-tone. WDER measures the proportion of diacritised reference words that the model does not reproduce exactly. BLEU and ChrF are computed using sacrebleu \citep{post-2018} with pinned tokenisation settings.

To prevent structural edits (insertions, deletions, substitutions) from inflating error metrics via misalignment, reference and output words are aligned on their unmarked base forms using Needleman-Wunsch alignment prior to evaluating diacritic accuracy \citep{needleman-wunsch-1970, aldallal-etal-2025}. Furthermore, references and outputs are normalised to a consistent Unicode form \citep{unicode-2025}. We also identified that 3.56\% of underdot diacritics in the YAD benchmark used non-standard Unicode codepoints. These were corrected prior to evaluation, and the correction patch is released alongside our code.

\subsection{Error typology and text fidelity}

While aggregate metrics like DER and CER quantify overall performance, they fail to characterise specific error types, such as whether a model struggles more with tone marks or underdots. To address this, we categorise every erroneous word into one of six main classes based on our base-word alignment:

\begin{itemize}
  \item \textbf{Diacritic errors (base form correct):}
  \begin{itemize}
    \item \emph{Tone-only:} one or more incorrect tone marks.
    \item \emph{Underdot-only:} one or more incorrect underdots.
    \item \emph{Mixed:} errors in both tone and underdots.
  \end{itemize}
  \item \textbf{Text-altering errors:}
  \begin{itemize}
    \item \emph{Lexical substitution:} the output word has a different base form than the reference.
    \item \emph{Deletion:} the reference word is absent from the output.
    \item \emph{Insertion:} the output includes a word not in the reference.
  \end{itemize}
\end{itemize}

We quantify text fidelity by calculating the ratio of lexical substitutions, deletions, and insertions to total errors, termed the text-altering rate. A lower ratio indicates higher fidelity, as errors are limited to diacritics. We report this metric for all seven models using a greedy decoding pipeline to ensure that the results reflect each model's natural tendencies.

\subsection{Environment}

The evaluation runs on a single NVIDIA T4 GPU. The software versions are pinned and are: Python 3.12.13, torch 2.10.0+cu128, transformers 4.57.6, sacrebleu 2.6.0.

\section{Results and Discussion}

\subsection{Overall comparison}

Table~\ref{tab:main} reports every metric for all seven models on the YAD test benchmark, under one decoding and scoring protocol (except for gpt-oss:120b), sorted by DER. We refer to the numbers in the table throughout the rest of this section.

\begin{table*}[t]
  \centering
  \small
  \caption{Comparison on the YAD test benchmark (3,330 sentences), corrected YAD test set, base-word Needleman-Wunsch alignment. Beam decoding (\texttt{num\_beams} = 5). Sorted by DER; best value per column in bold. * gpt-oss:120b is run one-shot at temperature 0, not beam search.}
  \label{tab:main}
  \begin{tabular}{@{}lrrrrrrrrr@{}}
    \toprule
    \textbf{Model} & \textbf{Params} & \textbf{CER} & \textbf{WER} & \textbf{DER} & \textbf{DER-tone} & \textbf{DER-und.} & \textbf{WDER} & \textbf{BLEU} & \textbf{ChrF} \\
    \midrule
    Yo-ByT5                & 299.6M        & \textbf{3.48} & 14.86          & \textbf{10.14} & \textbf{8.93} & \textbf{6.15} & 15.64          & 0.6841          & 0.8431 \\
    mT5-base               & 582.4M        & 4.38          & \textbf{14.47} & 10.28          & 9.27          & 7.62          & \textbf{13.92} & 0.7051          & \textbf{0.8506} \\
    omowe-T5 all-und       & 76.9M         & 13.36         & 20.33          & 14.76          & 14.12         & 13.11         & 16.68          & \textbf{0.7247} & 0.8295 \\
    omowe-T5 menyo         & 76.9M         & 35.18         & 43.51          & 20.46          & 19.56         & 21.24         & 19.94          & 0.5404          & 0.7562 \\
    gpt-oss:120b *         & $\approx$120B & 10.31         & 36.78          & 28.42          & 24.06         & 27.65         & 34.15          & 0.3801          & 0.6468 \\
    ByT5-small menyo       & 299.6M        & 30.84         & 44.98          & 40.86          & 38.96         & 35.38         & 45.71          & 0.3661          & 0.5829 \\
    mT5-small menyo        & 300.2M        & 35.33         & 62.19          & 44.53          & 42.71         & 49.17         & 44.62          & 0.2751          & 0.5448 \\
    \bottomrule
  \end{tabular}
\end{table*}

\subsection{Yo-ByT5 and mT5-base: parity at a fraction of the size}

Yo-ByT5 and mT5-base emerge as the leading ADR models for Yor\`ub\'a. On DER the two models are close enough that the gap is within the margin of error, as we confirmed with a resampling test where we redrew the test set 10,000 times. On the other metrics each model has an edge: Yo-ByT5 is better on CER, while mT5-base is better on word diacritic error rate and slightly better on WER. Yo-ByT5 reaches this level with 299.6M parameters against mT5-base's 582.4M, roughly half the size.

\subsection{Error analysis}

\begin{figure*}[t]
  \centering
  \includegraphics[width=\textwidth]{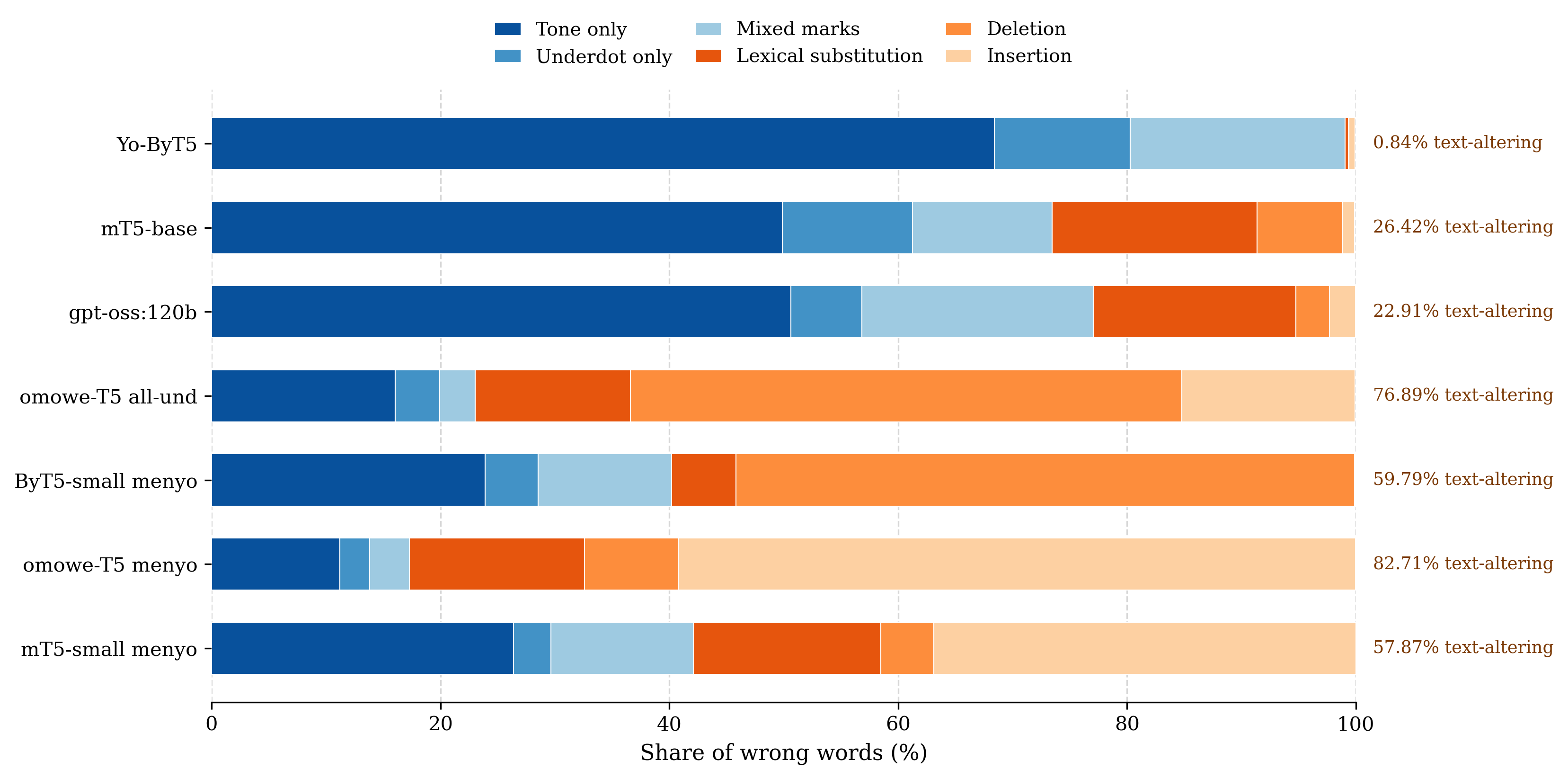}
  \caption{Word error typology across the seven evaluated models (corrected YAD test, greedy decoding). Blue hues: the three mark classes. Orange hues: the three text-altering classes. The text-altering share is stated beside each bar.}
  \label{fig:typology}
\end{figure*}

In addition to error rates, preserving input text fidelity is a critical requirement for ADR models. As shown in Figure~\ref{fig:typology}, Yo-ByT5 demonstrates superior fidelity, altering base input text in only 0.84\% of its errors. In contrast, mT5-base alters base text in 26.42\% of errors, gpt-oss:120b in 22.91\%, and other public baselines in 57.9\% to 82.7\% of errors.

An important finding is that byte-level architecture alone does not guarantee fidelity. The other publicly released byte-level model with the same underlying architecture as Yo-ByT5, ByT5-small menyo, alters text in 59.79\% of its errors, compared to 0.84\% for Yo-ByT5. This shows that training data and procedure remain paramount, and decide whether the inherent copying advantage of the byte-level architecture is realised.

We also find that tone-only errors outnumber underdot-only errors consistently across every model, indicating that the models struggle more with tone marks than with underdots. Furthermore, we discover that DER alone is not sufficient to evaluate a model, as one model, omowe-T5 menyo, scores a DER that puts it in the middle of the group, but its CER and WER are among the worst. This motivates reporting DER alongside WER and CER for a holistic view of a model's capabilities rather than DER alone.

\subsection{Open-weight language model baseline}

\begin{figure}[t]
  \centering
  \includegraphics[width=\columnwidth]{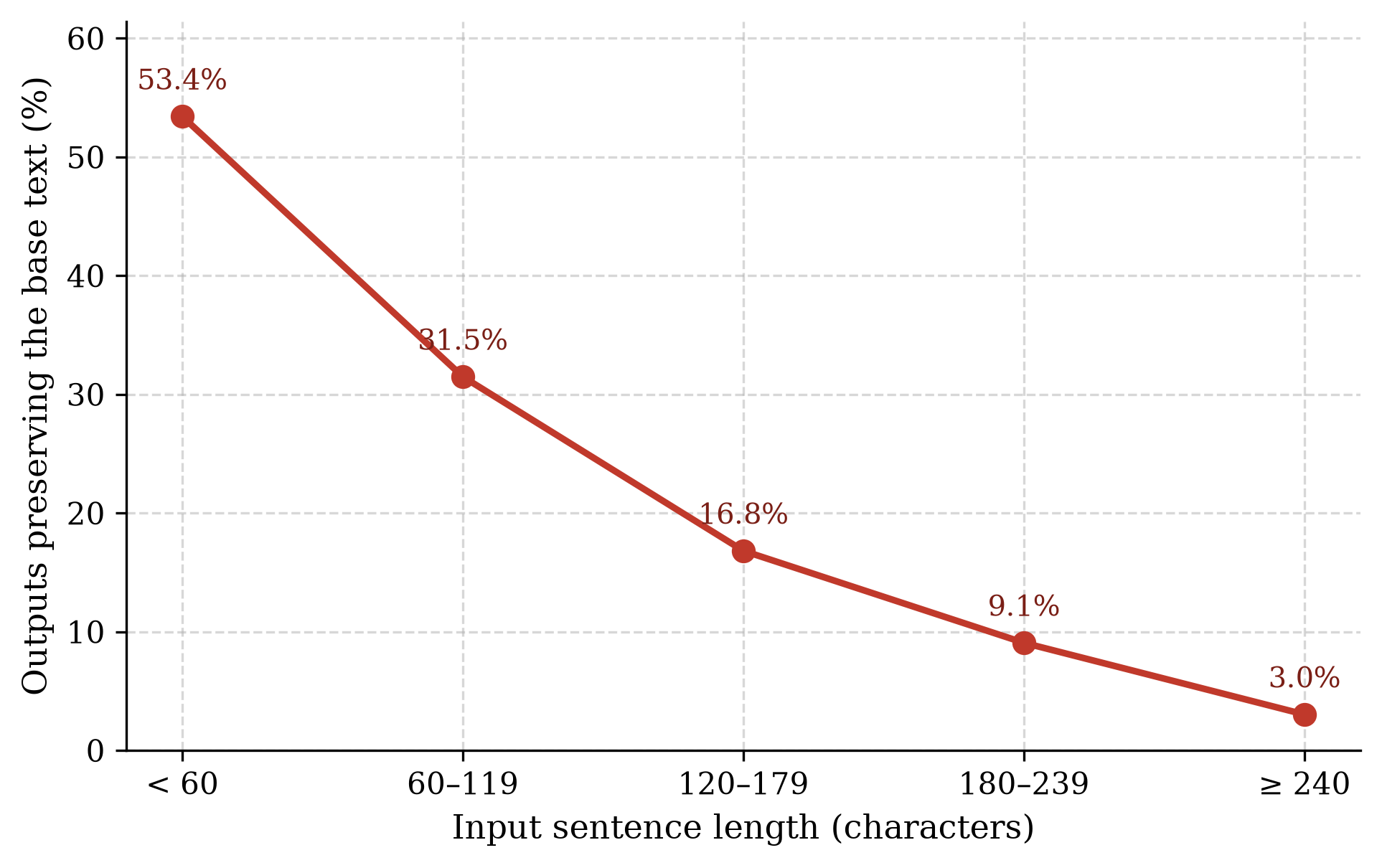}
  \caption{Base-text preservation by gpt-oss:120b against input sentence length (YAD test). The share of outputs that keep the base text falls from 53.4\% on the shortest sentences to 3.0\% on the longest.}
  \label{fig:fidelity}
\end{figure}

The 120B-parameter general large language model scores a DER of 28.42\% versus 10.14\% for Yo-ByT5. It ranks third among the seven models on CER and fifth on DER, indicating that it produces coherent output but unreliable diacritics. This corroborates that for the Yor\`ub\'a language, specialised fine-tuning still beats general-purpose language models on diacritic restoration \citep{nadas-diosan-2025}. Interestingly, text fidelity falls sharply with input length, from 53.4\% of sentences under 60 characters to 3.0\% of those with 240 or more. Figure~\ref{fig:fidelity} plots this decline across the five length bands.

\subsection{Decoding strategy}

The performance of the two most accurate models, Yo-ByT5 and mT5-base, remains steady across different decoding strategies, with greedy and beam decoding staying within 0.12 CER points for both models. However, for the weakest public models, beam search recovers up to 10.5 CER points. This suggests that a comparison that fixes one decoding strategy runs the risk of misranking the weaker models.

\section{Conclusion}

We introduce Yo-ByT5, a byte-level ADR model for Yor\`ub\'a, and evaluate it alongside five public diacritisers and an open-weight LLM under a unified benchmark protocol. Yo-ByT5 matches the accuracy of the top subword baseline while utilising half the parameters and exhibiting superior input text fidelity. Our analysis confirms that text fidelity is a key differentiator across models that is not captured by DER alone. All code, outputs, and dataset patches are released to support future research.

\section*{Limitations}

\paragraph{Strict word alignment.} The alignment algorithm used in the evaluation matches words one-to-one. Therefore, if a model splits or merges a word, the algorithm marks it as a substitution plus a gap rather than a simple change, which could slightly affect scores.

\paragraph{Narrow LLM evaluation.} Our LLM baseline was one model of one size with one prompt, and we report a single run. Future studies should perform further ablation studies, evaluating different models, using varying prompts, and running inference multiple times since LLMs sometimes give different outputs for the same input.

\paragraph{Training-register bias.} About 78\% of the training data is biblical text (36,243 of 46,313 sentences). Prior work has shown that over-reliance on biblical corpora can limit generalisation to colloquial text, contractions, and loan words \citep{orife-etal-2020}. Our evaluation uses the multi-domain YAD benchmark, so the reported scores are not measured on scripture, but a training corpus balanced across more domains may improve performance on informal text.

\paragraph{We need better benchmarks.} These limitations show why we need a larger, dedicated dataset for Yor\`ub\'a diacritic restoration. Future benchmarks should focus on diacritic restoration alone and cover a wider variety of contexts than the 3,330 sentences used in the YAD benchmark.

\section*{Acknowledgements}

This work was carried out within the HausaNLP Research Group. Research supported with Cloud TPUs from Google's TPU Research Cloud (TRC).

\bibliography{custom}

\end{document}